\documentclass[letterpaper, 10 pt, conference]{ieeeconf}  

\IEEEoverridecommandlockouts                              

\usepackage{arydshln}
\usepackage{hyperref}
\usepackage{multirow}

\usepackage{graphicx}
\usepackage{amsmath} 
\usepackage{amsfonts} 
\usepackage{balance}

\title{\LARGE \bf
EEG-to-Text Translation Using Deep Learning and LLMs*
}

\title{\LARGE \bf
RAG-based EEG-to-Text Translation Using Deep Learning and LLMs
}

\author{Enrico Collautti$^{1}$, Xiaopeng Mao$^{3}$, Luca Tonin$^{1,2}$, Stefano Tortora$^{1,2}$, Sadasivan Puthusserypady$^{3}$  
\thanks{This work has been submitted to the IEEE for possible publication. Copyright may be transferred without notice, after which this version may no longer be accessible.}
\thanks{$^{1}$IAS-LAB, Department of Information Engineering,
        University of Padova, 35122 Padova (PD), Italy
        {\tt\small enrico.collautti@studenti.unipd.it}
        {\tt\small \{luca.tonin, stefano.tortora\}@unipd.it}}
\thanks{$^{2}$Padova Neuroscience Center, 35131 Padova (PD), Italy.}%
\thanks{$^{3}$Department of Health Technology, Technical University of Denmark, 2800 Kgs. Lyngby, Denmark 
{\tt\small \{xiama, sapu\}@dtu.dk}
}
}
\begin{document}

\maketitle
\thispagestyle{empty}
\pagestyle{empty}


\begin{abstract}

The decoding of linguistic information from electroencephalography (EEG) signals remains an extremely challenging problem in brain-computer interface (BCI) research. In particular, sentence-level decoding from EEG is difficult due to the low signal-to-noise ratio of these recordings. Previous studies tackling this problem have typically failed to surpass random baseline performance unless teacher forcing is used during the inference phase. 
In this work, we propose a retrieval-augmented generation (RAG)-based sentence-level EEG-to-text decoding pipeline that combines an EEG encoder aligned with semantic sentence embeddings, a vector retrieval stage, and a large language model (LLM) to refine retrieved sentences into coherent output. Experiments are conducted on the Zurich Cognitive Language Processing Corpus (ZuCo) dataset, which contains single-trial EEG recordings collected during silent reading. To evaluate whether the system extracts meaningful information from these EEG signals, the results are compared with a random baseline.
In nine subjects, the proposed pipeline outperforms the random baseline, achieving a mean cosine similarity of $0.181 \pm 0.022$ compared to $0.139 \pm 0.029$ for the baseline, corresponding to a relative improvement of $30.45\%$. Statistical analysis further confirms that this improvement is significant, following a strict evaluation workflow where inference is performed without access to ground-truth labels. 

\end{abstract}


\section{INTRODUCTION}

Brain-computer interfaces (BCIs) aim to translate neural activity into external control signals, providing alternative communication pathways for individuals with severe motor impairments, enabling users to express intentions without relying on muscular activity. Recent advances in invasive neural recording technologies, such as electrocorticography (ECoG) and intracortical implants, have demonstrated promising results in decoding speech and linguistic information directly from brain activity~\cite{willett_high-performance_2023, card_accurate_2024}.
However, invasive BCIs require neurosurgical implantation and are therefore limited to specific clinical contexts. For broader and safer deployment, non-invasive approaches such as electroencephalography (EEG) remain highly attractive due to their safety, portability, and relatively low cost. BCIs based on EEG have been developed for a variety of tasks, including motor control, environmental interaction, and communication~\cite{tonin_noninvasive_2021}.
Despite these advantages, decoding natural language from EEG signals remains extremely challenging. EEG recordings suffer from low spatial resolution, high noise levels, and strong inter-subject variability. As a result, many early systems relied on restricted vocabularies or character-level decoding strategies, which significantly limit fluency and practicality. Consequently, reliable sentence-level EEG-to-text decoding remains an open challenge.

Recent work has also raised concerns about the evaluation methodologies commonly used in EEG-to-text research. Jo et al.~\cite{jo_evaluating_2025} demonstrate that several neural decoding models report strong performance despite relying on questionable evaluation practices. In particular, some systems employ teacher forcing during inference, meaning that ground-truth tokens are partially provided to the decoder during testing. This practice artificially simplifies the generation task and can significantly inflate reported metrics. In addition, the authors highlight that many studies lack meaningful comparisons against random or noise-based baselines. Through a noise-based analysis, they show that some EEG-to-text models achieve comparable performance even when the input EEG signals are replaced with noise, suggesting that the models may rely on dataset artifacts or memorization rather than genuinely decoding neural information. These findings emphasize the importance of rigorous evaluation protocols and proper baselines when assessing brain-to-text systems.

Recent approaches have explored the integration of deep learning and large language models (LLMs) to improve EEG-to-text decoding. For example, Mishra et al.~\cite{mishra_thought2text_2025} propose \textit{Thought2Text}, a three-stage pipeline that combines an EEG encoder with multimodal alignment and instruction-tuned LLMs. Their method trains a multichannel convolutional EEG encoder based on ChannelNet~\cite{palazzo_decoding_2021} to align EEG embeddings with pretrained contrastive language-image pre-training (CLIP)~\cite{radford_learning_2021} image embeddings while simultaneously predicting salient object labels. The resulting multimodal representation is then projected onto the token embedding space of an LLM, enabling the generation of textual descriptions conditioned on EEG-derived signals. While this multimodal alignment strategy demonstrates competitive performance, it relies on visual stimuli and image representations, which restrict the expressive freedom of natural language. Furthermore, projecting EEG embeddings directly onto the token embedding space of a language model may introduce information loss due to the compression required by the projection layer.

In this work, a sentence-level embedding strategy is adopted to support more natural and flexible communication. A modular EEG-to-text framework is proposed in which each component can be independently replaced or improved, providing a flexible foundation for future developments in artificial intelligence and neuroengineering. The pipeline learns EEG sentence embeddings aligned with pretrained language representations, enabling semantic retrieval in a vector space without directly projecting neural embeddings onto the language model’s internal representation. Retrieved candidate sentences are subsequently refined by an LLM performing a natural language processing task.
In addition, the proposed framework introduces a rigorous evaluation protocol designed to avoid the methodological pitfalls highlighted in recent literature. The system is evaluated without any form of teacher forcing during inference, ensuring that the decoder does not receive ground-truth information at test time. Performance is compared against a random baseline following the evaluation concerns raised by Jo et al.~\cite{jo_evaluating_2025}. However, beyond simple baseline comparison, this work further introduces statistical analysis to determine whether the observed differences between real EEG decoding and random baselines are statistically significant. This additional validation step provides a stronger and more reliable assessment of whether the decoded information genuinely originates from neural signals rather than dataset artifacts or chance effects.
\vspace{-0.14cm}

\section{METHODS}

\subsection{Dataset}

The Zurich Cognitive Language Processing Corpus (ZuCo) 1.0 and 2.0 are datasets that provide EEG recordings collected during silent reading of English sentences~\cite{hollenstein_zuco_2018, hollenstein_zuco_2020}. The ZuCo datasets are well-suited for brain-to-text research because they (i) use a silent reading paradigm that avoids speech-related muscle artifacts, (ii) rely on EEG rather than invasive recording, and (iii) provide labeled EEG-text pairs that make it possible to train supervised deep learning models for mapping neural activity to linguistic representations. The datasets include multiple reading tasks: \textit{Natural Reading} (NR), in which participants read sentences normally; \textit{Sentiment Reading} (SR), where participants read sentences and subsequently evaluate their sentiment; and \textit{Task-Specific Reading} (TSR), which involves an active language comprehension exercise. In addition, the datasets provide eye-tracking annotations that allow precise alignment between textual stimuli and neural signals.

In this work, subject-dependent sentence-level decoding is performed across nine ZuCo~1.0 participants. Three out of the twelve subjects were excluded due to incomplete recordings that resulted in substantially fewer usable sentence trials. For each included subject, the NR and SR tasks were merged after preprocessing, yielding approximately 700 sentence-level EEG-text pairs per subject. Models are trained and evaluated independently for each subject in order to assess robustness across individuals.

\subsection{Main Pipeline}

\begin{figure*}[t]
    \centering
    \includegraphics[width=0.8\textwidth]{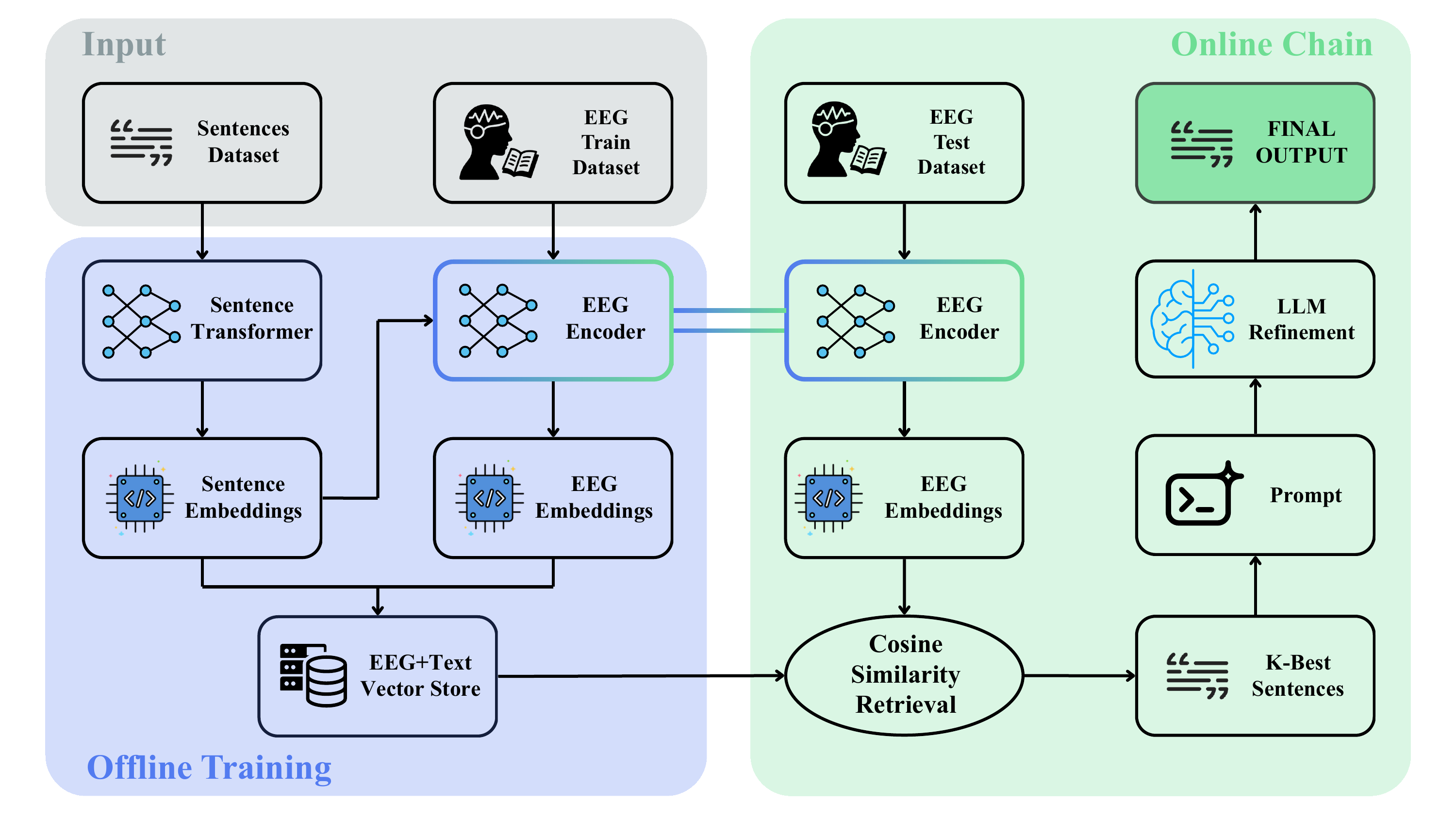}
    \caption{Overview of the proposed EEG-to-text decoding pipeline. The framework consists of an offline training stage for learning EEG sentence embeddings and an online inference stage that performs similarity retrieval and language model refinement.}
    \label{fig:pipeline}
    \vspace{-0.5cm}
\end{figure*}

The overall architecture of the proposed EEG-to-text decoding framework is illustrated in Fig.~\ref{fig:pipeline}. The pipeline is organized into two main stages: an \textit{offline training phase} and an \textit{online inference phase}. 

The individual components of the pipeline are described in detail in the following subsections.

\subsection{EEG Preprocessing}

EEG signals in ZuCo 1.0 were originally acquired using a 128-channel Geodesic Hydrocel system sampled at 500~Hz with hardware band-pass filtering of 0.1 to 100~Hz~\cite{hollenstein_zuco_2018}. The dataset provides artifact-cleaned signals produced using the Automagic preprocessing pipeline~\cite{pedroni_automagic_2019}. This pipeline includes bad-channel detection and interpolation, high-pass and notch filtering, eye-artifact regression, independent component analysis (ICA) with automatic artifact component rejection using Multiple Artifact Rejection Algorithm (MARA), and average re-referencing~\cite{delorme_eeglab_2004,winkler_automatic_2011}.

Sentence-level EEG segments are extracted using fixation-aligned timing information provided by the dataset. Each segment corresponds to the EEG activity recorded during the reading of a specific sentence. Since sentence durations vary across trials, all EEG segments among the same subject are symmetrically zero-padded to a common temporal length, enabling batch processing with convolutional neural networks (CNNs).

\subsection{Sentence Embedding Targets}

Each ground-truth sentence is mapped to a contextual semantic representation using a pretrained sentence embedding model. Specifically, the \texttt{all-mpnet-base-v2} model from the Sentence Transformers library is used to compute 768-dimensional sentence embeddings~\cite{noauthor_sentence-transformersall-mpnet-base-v2_2024}. These embeddings serve as the target representations that the EEG encoder is trained to approximate.

\subsection{EEG Encoder and Embedding Alignment}

A convolutional EEG encoder is trained to map a multi-channel EEG segment $\mathbf{X}\in\mathbb{R}^{C\times T}$ (channels $\times$ time samples) to a 768-dimensional embedding $\mathbf{z}_{\text{EEG}}\in\mathbb{R}^{768}$. The architecture is inspired by EEGChannelNet-style designs \cite{palazzo_decoding_2021}. It combines (i) temporal depthwise convolutions with multiple dilation factors to capture multi-scale temporal dynamics, (ii) spatial convolutions to model inter-channel relationships, and (iii) residual blocks to improve training stability.
A global average pooling layer aggregates temporal features, followed by a linear projection that maps the representation to the 768-dimensional sentence embedding space in order to align it with the precomputed text embeddings.

The model is trained using a cosine similarity loss that encourages alignment between EEG and text embeddings:

\begin{equation}
\mathcal{L} = 1 - \frac{\mathbf{z}_{\text{EEG}} \cdot \mathbf{z}_{\text{TXT}}}
{\|\mathbf{z}_{\text{EEG}}\| \|\mathbf{z}_{\text{TXT}}\|},
\label{loss function}
\end{equation}
where $\mathbf{z}_{\text{TXT}}$ denotes the target sentence embedding.

For each subject, the dataset of approximately 700 sentences is divided into 650 training samples and 50 test samples in order to evaluate generalization under limited data conditions. The 50 test sentences are identical across subjects, allowing semantically consistent comparisons between participants. A preliminary hyperparameter search is conducted using a validation set of 50 sentences drawn from the training portion. The objective of this search is to identify a single hyperparameter configuration that can be applied consistently across all subjects, thereby testing the generalization capability of the encoder architecture.

The explored optimization strategies include dropout regularization, L2 weight decay, LeakyReLU activations, and a cosine annealing learning-rate scheduler. The grid search is executed using three different random seeds to reduce sensitivity to initialization effects. The hyperparameter configuration that demonstrated the most stable and consistent performance across all seeds is selected. After selecting the best configuration, the model is retrained on the full training set (without a validation split) to produce the final EEG embeddings for each subject.

\subsection{Vector Indexing and Retrieval-Augmented Generation}

After training, EEG embeddings corresponding to all training sentences are indexed in a Facebook AI Similarity Search (FAISS)~\cite{johnson_billion-scale_2021} vector store to enable efficient nearest-neighbor retrieval using cosine similarity. Importantly, the vector store is constructed exclusively from training sentences, and all test sentences are strictly excluded. As a result, during inference, the retrieval stage cannot return the ground-truth sentence associated with the test EEG signal, preventing any form of data circularity.
During retrieval, a test EEG segment is encoded into an embedding $\mathbf{z}_{\text{EEG}}$, and the top-$k$ nearest sentences are retrieved from the vector store (with $k=3$).

Inference is accomplished via a retrieval-augmented generation (RAG), where the retrieved sentences are provided to an LLM, which generates a single fluent output sentence that captures their shared semantic content. In this work, the \texttt{meta-llama/Meta-Llama-3-8B} model~\cite{grattafiori_llama_2024} is used through HuggingFace. The model is prompted with a constrained instruction designed to limit hallucination and enforce semantic preservation:

\textit{``You are given three similar sentences. Your task is to generate one single, fluent, grammatically correct sentence that captures the same meaning, without adding or removing information. Only output the corrected sentence.''}

\subsection{Random Baseline and Statistical Analysis}

To assess whether the proposed EEG-to-text decoding pipeline performs above chance level, a random baseline is constructed and compared with the real decoding results.

The random baseline is generated by shuffling the EEG signals associated with each sentence. Specifically, the original EEG data for each trial (channels $\times$ time samples) is randomly permuted along the temporal dimension to produce synthetic signals that preserve the physiological amplitude distribution of EEG activity while destroying its temporal structure. As a result, the mean and standard deviation of the signals remain unchanged, while the temporal structure is disrupted, thereby altering spectral-temporal patterns that may encode linguistic information. The shuffled signals are then randomly assigned to the true sentence labels of the test set and processed through the same decoding pipeline used for real EEG data. To account for variability in the randomization process, this procedure is repeated across multiple random seeds (here $n=10$).

For evaluation, both the generated sentence and the corresponding ground-truth sentence are embedded, and the cosine similarity between the two embeddings is used as a measure of semantic alignment.

To determine whether decoding performance exceeds chance level, statistical comparisons between real decoding results and the random baseline are conducted at both the subject-specific level and the whole-dataset level.

At the whole-dataset level, analysis is performed across subjects. For each subject, cosine similarity scores obtained from the random baseline are first averaged across the different random seeds. The mean cosine similarity is then computed separately for the real decoding outputs and the random baseline, producing one paired observation per subject (real mean vs.\ random mean). These paired values are compared using a Wilcoxon signed-rank test with a one-sided alternative hypothesis ($\text{real} > \text{random}$)~\cite{gibbons_nonparametric_2025}, reflecting the hypothesis that real EEG decoding should outperform the random baseline.

At the subject-specific level, decoding performance is evaluated individually for each participant. For each subject, the cosine similarity scores of the 50 decoded sentences are compared with the corresponding random baseline scores obtained by averaging across random seeds for each sentence. The comparison between real and random cosine similarities is again evaluated using a Wilcoxon signed-rank test with a one-sided alternative hypothesis.

Since these statistical tests are conducted separately for multiple subjects, the resulting $p$-values are corrected for multiple comparisons using the Benjamini-Hochberg False Discovery Rate (FDR) procedure.

\section{RESULTS}

\subsection{EEG Encoder Training}


During the training phase, the optimized configuration shows substantially improved generalization compared to a direct adaptation of the baseline EEGChannelNet-style encoder. Training and validation losses decrease together throughout the optimization process, with no evident divergence that would indicate overfitting. This behavior is consistent across subjects, with similar convergence patterns observed for all participants. Final validation losses across subjects remain in the $0.60$--$0.70$ range. The cosine loss, as defined in Eq.~\ref{loss function}, ranges between $0$ and $2$, and these values are therefore consistent with the noisy nature of single-trial EEG signals and the limited number of training examples available for each subject.

\subsection{Final Output Examples}

\begin{table}[t]
\centering
\small
\caption{Examples of decoded sentences produced by the pipeline.}
\label{tab:good_examples}

\renewcommand{\arraystretch}{1.4}

\begin{tabular}{|p{8.0cm}|}
\hline

\textbf{Subject:} ZMG

\textbf{Target Sentence:} ... by the time it's done with us, Mira Nair's new \underline{movie has its audience giddy with the delight of discovery}, of having been immersed in a foreign culture only to find that human nature is pretty much the same all over.

\textbf{Final Output:} \underline{A beautifully crafted, emotionally resonant} \underline{film} that should appeal to \underline{audiences} looking for something \underline{original and thought-provoking}.

\textbf{Cosine Similarity Score:} 0.560

\\ \hline

\textbf{Subject:} ZJN

\textbf{Target Sentence:} Huxley \underline{was born} in Godalming, Surrey, \underline{England}.

\textbf{Final Output:} He \underline{was born} in Litchfield, \underline{England}, where he attended school before moving on to study medicine at Edinburgh University, Scotland.

\textbf{Cosine Similarity Score:} 0.551

\\ \hline

\textbf{Subject:} ZMG

\textbf{Target Sentence:} \underline{The film is so bad} it doesn't improve upon the experience of staring at a blank screen.

\textbf{Final Output:} A very long \underline{film} with \underline{dull} parts, excessive emphasis on food preparation and snow hut building, and a leading actress who \underline{ruins it all} with her utter \underline{inability to act}.

\textbf{Cosine Similarity Score:} 0.501

\\ \hline

\textbf{Subject:} ZDM

\textbf{Target Sentence:} The simplicity of The Way Home has few equals this side of Aesop.

\textbf{Final Output:} Bogdanovitch's film is an interesting look at the Hearsts' history, but the climax of the story gets lost in its own excesses.

\textbf{Cosine Similarity Score:} 0.101

\\ \hline

\end{tabular}
\end{table}

As a qualitative evaluation of the proposed pipeline, Table~\ref{tab:good_examples} reports some representative decoding results across different subjects. For each example, the target sentence, the generated output, and the cosine similarity score between their sentence embeddings are shown. 
The underlined segments highlight portions of the generated output that are semantically similar to the target sentence. These correspondences are not necessarily exact word matches, but rather expressions that convey a similar meaning or concept. This reflects the goal of the proposed pipeline, which is to capture semantic information from EEG signals rather than reproduce the exact lexical structure of the original sentence.
In the successful examples, the system correctly identifies the main topic of the sentence, such as references to a \textit{movie}, and captures the associated sentiment or overall context (e.g., describing a film in a positive or negative way). This suggests that the decoding pipeline is able to extract coarse semantic cues from neural activity. 
For completeness, a failure case is also shown in the last example of the table. In this example, the generated sentence is unrelated to the target content, illustrating a hallucination produced by the retrieval and generation stages.

\subsection{Random Baseline and Statistical Analysis}

\begin{table}[t]
\centering
\small
\setlength{\tabcolsep}{3pt} 
\renewcommand{\arraystretch}{1.02} 

\caption{Subject-specific and whole-dataset analysis}
\label{tab:results_analysis}

\begin{tabular}{c cc c c c}
\hline
\\[-8pt]

\multirow{2}{*}{\textbf{Subj.}} &
\textbf{Real CosSim} &
\textbf{Random CosSim} &
\multirow{2}{*}{\textbf{$\Delta$}} &
\multirow{2}{*}{\textbf{$\Delta$\%}} &
\multirow{2}{*}{\textbf{$p_w$}} \\

& \textbf{$(\mu \pm \sigma)$} & \textbf{$(\mu \pm \sigma)$} & & & \\

\\[-8pt]
\hline
\\[-8pt]

ZAB & $0.214\pm0.111$ & $0.147\pm0.152$ & 0.067 & 46.02 & \textbf{0.012} \\
ZDM & $0.148\pm0.128$ & $0.141\pm0.136$ & 0.008 & 5.52 & 0.417 \\
ZGW & $0.195\pm0.159$ & $0.115\pm0.140$ & 0.080 & 70.22 & \textbf{0.009} \\
ZJM & $0.187\pm0.139$ & $0.174\pm0.154$ & 0.013 & 7.51 & 0.417 \\
ZJN & $0.158\pm0.133$ & $0.105\pm0.097$ & 0.053 & 50.50 & \textbf{0.030} \\
ZKB & $0.165\pm0.128$ & $0.172\pm0.146$ & -0.008 & -4.43 & 0.594 \\
ZKH & $0.208\pm0.153$ & $0.096\pm0.111$ & 0.112 & 116.11 & \textbf{0.009} \\
ZKW & $0.171\pm0.129$ & $0.165\pm0.129$ & 0.006 & 3.43 & 0.417 \\
ZMG & $0.185\pm0.158$ & $0.136\pm0.101$ & 0.049 & 36.10 & \underline{0.056} \\

\\[-8pt]
\hline\hline
\\[-8pt]

ALL & $0.181\pm0.022$ & $0.139\pm0.029$ & 0.042 & 30.45 & \textbf{0.0059} \\

\\[-8pt]
\hline
\end{tabular}

\vspace{-0.5cm}
\end{table}

\begin{figure}[t]
\centering
\includegraphics[width=\columnwidth]{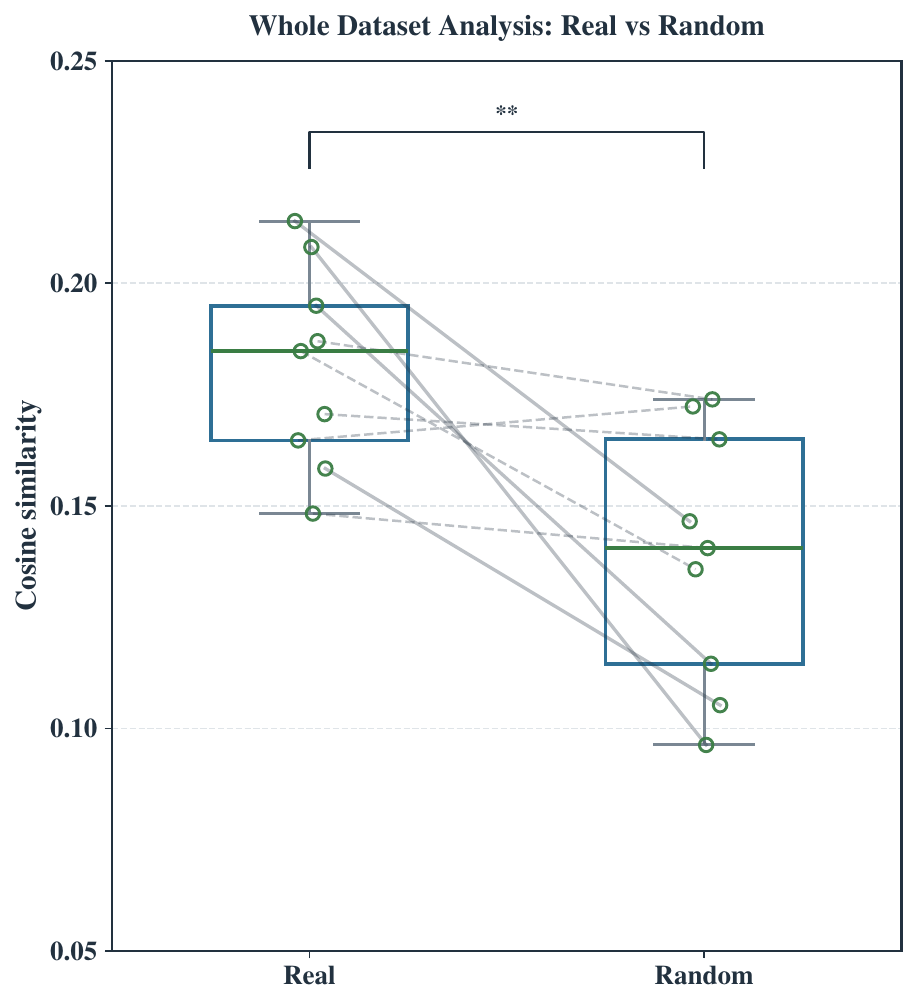}
\caption{Whole-dataset comparison between real and random cosine similarity distributions. Boxplots show the interquartile range (IQR), median, and whiskers. Each subject is connected across conditions by a line, which is solid for statistically significant differences ($p < 0.05$) and dashed otherwise. The double asterisk ($^{**}$) indicates a highly significant overall difference ($p < 0.01$).}
\label{fig:dataset_real_vs_random}
\vspace{-0.5cm}
\end{figure}

Table~\ref{tab:results_analysis} reports, respectively, the comparison between real decoding results and the random baseline at the subject-level and the overall behavior of the system across participants. Cosine similarity is computed between the embedding of the generated sentence and that of the ground-truth sentence. The tables show the mean and standard deviation of cosine similarity scores obtained from the real decoding pipeline and from the random baseline. The columns $\Delta$ and $\Delta\%$ indicate the absolute and relative improvements of the real decoding results over the random baseline. The column $p_w$ reports the $p$-values obtained from the Wilcoxon signed-rank test, following the evaluation procedure described in the Methods section. Values below $0.05$ are reported in \textbf{bold}, indicating statistically significant results. The same statistical comparison is further illustrated by the boxplots in Fig.~\ref{fig:dataset_real_vs_random}.

At the subject-specific level, most subjects exhibit higher cosine similarity values for the real decoding pipeline compared with the random baseline. Statistically significant improvements are observed for subjects ZAB, ZGW, ZJN, and ZKH. Subject ZMG also shows a positive improvement but does not reach the significance threshold and is therefore reported as a borderline case (underlined). In particular, the $p$-value for ZMG exceeds the $0.05$ threshold after applying the Benjamini/Hochberg False Discovery Rate (FDR) correction.
At the whole-dataset level, on average, the real decoding pipeline achieves higher cosine similarity scores than the random baseline across subjects, a result that is further supported by a highly significant $p$-value ($p < 0.01$).

Overall, the whole-dataset results confirm that the proposed decoding pipeline produces outputs that are significantly more semantically aligned with the target sentences than those obtained from the random baseline. Although the absolute cosine similarity values remain moderate due to the intrinsic noise of single-trial EEG signals and the difficulty of sentence-level decoding, the statistical results indicate that the system captures meaningful semantic information from neural activity beyond chance-level performance.

\section{DISCUSSION}

To the best of our knowledge, this work presents the first sentence-level EEG-to-text decoding system that demonstrates performance significantly above a random baseline through statistical analysis. Although the cosine similarity scores remain moderate, the results are impressive given the difficulty of the task and the constraints of the experimental setup.

In particular, inference is performed without access to ground-truth sentences or labels during generation. The system therefore avoids optimistic estimates that may arise from teacher forcing strategies often used in sequence generation models, operating in a strictly constrained setting.

Another important challenge concerns the evaluation of EEG-to-text systems. Many previous studies rely on metrics such as (Bilingual Evaluation Understudy) (BLEU) or Recall-Oriented Understudy for Gisting Evaluation (ROUGE), which measure performance based on exact word or $n$-gram overlap between generated and reference sentences. While appropriate for tasks such as machine translation, these metrics are less suitable for EEG-based language decoding, where recovering the exact word sequence originally processed by the brain is extremely difficult~\cite{blagec_global_2022}.

In this context, the primary objective is to recover the semantic content of the intended message rather than to reproduce identical wording. For this reason, this work evaluates decoding performance using cosine similarity between sentence embeddings, which is widely adopted as a measure of semantic similarity in natural language processing (NLP). In addition, a statistical analysis comparing the real decoding pipeline with a randomized baseline is introduced to assess whether the observed improvements exceed chance-level performance. This evaluation criterion resembles a real-world scenario and is therefore more robust than using a white noise baseline~\cite{jo_evaluating_2025}. Although cosine similarity and statistical testing are well-established tools in machine learning (ML) and NLP, their combined use for evaluating EEG-to-text decoding pipelines has not been systematically explored in prior work.

Some limitations of the current approach must nevertheless be acknowledged. First, the ZuCo datasets provide only single-trial recordings for each sentence, meaning that each sentence is associated with a single EEG instance. From an ML perspective, this severely limits the amount of training data available for learning robust neural representations. In contrast to many deep learning applications where multiple samples per class are available, the present setup requires the model to generalize from highly sparse examples.
The goal of the decoding process is therefore to capture the general meaning of a sentence rather than to reproduce its exact wording. Hence, the finer-grained linguistic nuances are not captured, and the conventional NLP metrics are inapplicable. Nonetheless, this design choice aligns with the inherent variability and noise present in non-invasive EEG signals.

Despite these limitations, the modular structure of the proposed pipeline offers several advantages for future research. The architecture separates EEG representation learning, retrieval in the embedding space, and linguistic refinement through a language model. This separation allows each component to be independently improved or replaced without redesigning the entire system.
Among the components, the EEG encoder remains the most challenging part of the pipeline. Learning a meaningful embedding space from EEG signals is difficult, as the model must organize representations so that semantically related sentences are mapped to nearby points while unrelated sentences remain distant. In practice, the learned EEG embedding space often exhibits reduced dispersion, making retrieval more difficult. 

Future work could explore contrastive learning strategies to explicitly enforce separation between embeddings of different sentences and improve the structure of the learned representation space~\cite{khosla_supervised_2020}.
Additional improvements may also be achieved by exploring alternative sentence embedding models or more efficient LLMs for the refinement stage. In particular, prompt design and prompt engineering could play a role in improving the final linguistic output by guiding the language model to better integrate the retrieved sentences.

Finally, while the present work focuses on open-domain sentence decoding, future datasets designed specifically for assistive applications could significantly improve performance. For instance, a dataset tailored for impaired subjects with a limited vocabulary of functional commands (e.g., ``go to the bathroom'', ``bring the cup'', or ``turn on the light'') and multiple training trials per command would allow the system to learn more robust representations. In such scenarios, EEG decoding combined with modern vision-language-action models could potentially enable a natural human-robot interaction for assistive technologies capable of helping users in everyday life~\cite{sapkota_vision-language-action_2026}.

\section{CONCLUSION}

This study introduced a modular, RAG-based framework for brain-to-text decoding, successfully reconstructing semantic sentence content from non-invasive EEG signals. By aligning neural embeddings with high-dimensional language representations and using vector similarity retrieval, our pipeline demonstrates that EEG contains sufficient signal-to-noise to support sentence-level decoding significantly above chance level. While the inherent noise and single-trial nature of current datasets, such as ZuCo, limit exact word-for-word reconstruction, our results confirm that semantic intent is preserved and recoverable. Future advancements in discriminative embedding spaces and refined prompt engineering for the LLM stage will further narrow the gap between neural intent and linguistic output. Ultimately, the modularity of this RAG-based approach provides a scalable foundation for next-generation assistive BCIs, offering a promising path toward high-fidelity communication for speech-impaired users.

\addtolength{\textheight}{-12cm}   




\section*{\small ACKNOWLEDGMENT}
\small
\noindent Generative AI tools were used to support English grammar and code debugging. All content, decisions and results remain the responsibility of the authors.


\bibliographystyle{IEEEtran}
\balance
\bibliography{references_clean}

\end{document}